\documentclass[10pt,twoside]{article}

\usepackage{times}
\usepackage[utf8]{inputenc}
\usepackage[T1]{fontenc}
\usepackage{graphicx}

\usepackage{siunitx}
\usepackage{hyperref}
\usepackage{cleveref}
\usepackage{coria-taln2025}
\usepackage[french]{babel} 

\title{Vers une évaluation rigoureuse des systèmes RAG : le défi de la due diligence}

\author{Grégoire Martinon\up{1} \quad Alexandra Lorenzo de Brionne\up{2} \quad Jérôme Bohard\up{1} \quad Antoine Lojou\up{1} \quad Damien Hervault\up{1} \quad Nicolas Brunel\up{1, 3}\\
  {\small
    (1) Capgemini Invent France, 147 Quai du Président Roosevelt, 92130, Issy-les-Moulinaeux, France \\ 
    (2) DiaDeep, 35 Rue Louis Guérin, 69100, Villeurbanne, France \\
    (3) LaMME, ENSIIE, Université Paris-Saclay, 3 rue Joliot Curie, Bâtiment Breguet, 91190 Gif-sur-Yvette, France \\ 
    \texttt{
      gregoire.martinon@capgemini.com, a.lorenzo@diadeep.com jerome.bohard@capgemini.com, antoine.lojou@capgemini.com, damien.hervault@capgemini.com, nicolas.brunel@capgemini.com \\ 
}}}

\begin{document}
\maketitle

\resume{
L'IA générative se déploie dans des secteurs à haut risque comme la santé et la finance. L'architecture RAG (Retrieval Augmented Generation), qui combine modèles de langage (LLM) et moteurs de recherche, se distingue par sa capacité à générer des réponses à partir de corpus documentaires. Cependant, la fiabilité de ces systèmes en contextes critiques demeure préoccupante, notamment avec des hallucinations persistantes. Cette étude évalue un système RAG déployé chez un fonds d'investissement pour assister les due diligence. Nous proposons un protocole d'évaluation robuste combinant annotations humaines et LLM-Juge pour qualifier les défaillances du système, comme les hallucinations, les hors-sujets, les citations défaillantes ou les abstentions. Inspirés par la méthode Prediction Powered Inference (PPI), nous obtenons des mesures de performance robustes avec garanties statistiques. Nous fournissons le jeu de données complet. Nos contributions visent à améliorer la fiabilité et la scalabilité des protocoles d'évaluations de systèmes RAG en contexte industriel.
}

\abstract{Towards a rigorous evaluation of RAG systems: the challenge of due diligence}{
The rise of generative AI, has driven significant advancements in high-risk sectors like healthcare and finance. The Retrieval-Augmented Generation (RAG) architecture, combining language models (LLMs) with search engines, is particularly notable for its ability to generate responses from document corpora. Despite its potential, the reliability of RAG systems in critical contexts remains a concern, with issues such as hallucinations persisting. This study evaluates a RAG system used in due diligence for an investment fund. We propose a robust evaluation protocol combining human annotations and LLM-Judge annotations to identify system failures, like hallucinations, off-topic, failed citations, and abstentions. Inspired by the Prediction Powered Inference (PPI) method, we achieve precise performance measurements with statistical guarantees. We provide a comprehensive dataset for further analysis. Our contributions aim to enhance the reliability and scalability of RAG systems evaluation protocols in industrial applications.
}

\motsClefs
  {LLM, RAG, hallucinations, annotations, LLM-Juge, due diligence}
  {LLM, RAG, hallucinations, annotations, LLM-as-Judge, due diligence}

\acceptedArticle[accepted]{EvalLLM2025 : Atelier sur l'évaluation des modèles génératifs (LLM) et challenge} 

\section{Introduction}

Depuis l’apparition de ChatGPT, l’usage de l’IA générative s’est rapidement étendu à des secteurs sensibles comme la santé, la finance ou la défense. Parmi les applications notables, les systèmes RAG (Retrieval-Augmented Generation), combinant LLM (Large Language Model) et moteur de recherche, se distinguent par leur capacité à générer des réponses fondées sur un corpus documentaire, avec citations à l’appui.

Ces systèmes sont déjà largement adoptés dans l’industrie, notamment pour interroger des documentations d'entreprises complexes ou analyser des archives dans le cadre de fusions-acquisitions. Mais dans des contextes critiques, leur fiabilité est un enjeu crucial \cite{weidinger2025toward, zhou2024trustworthiness}. Malgré leurs promesses, les RAGs ne garantissent pas l’absence d’hallucinations \cite{magesh2024hallucination}, et leurs performances varient fortement selon le domaine métier. Cela impose des protocoles d’évaluation adaptés, en cohérence avec des exigences réglementaires croissantes, comme celles de l’AI Act.

Deux approches d'évaluation dominent : l’annotation humaine, précise mais coûteuse, et l’évaluation par LLM-Juge, scalable mais parfois biaisée. Des méthodes hybrides comme PPI (Prediction Powered Inference) \cite{doi:10.1126/science.adi6000, angelopoulos2023ppi++, boyeau2024autoeval} ou ASI (Active Statistical Inference) \cite{10.5555/3692070.3694680, gligoric2024can} proposent de les combiner pour une évaluation rigoureuse et économiquement viable. Dans cet article, nous exploitons ces méthodes pour évaluer un système RAG utilisé lors de due diligence. Nous développons un protocole approfondi, inspiré de PPI, pour quantifier les points de défaillance d’un système industriel \footnote{De récents travaux proposent de travailler à la maille du fait atomique généré dans le texte \cite{min2023factscore, scire2024truth}. Cependant, nous avons constaté que cette maille était extrêmement laborieuse à extraire pour les annotateurs humains et difficilement automatisable pour les LLMs, avec des taux d'oublis de l’ordre de 50\%. Nous faisons donc le choix pragmatique de travailler à l’échelle de la phrase pour quantifier et localiser les hallucinations.}.

Nos contributions principales sont les suivantes :

\begin{itemize}

    \item Un protocole d'évaluation détaillé, par domaine métier, des performances d’un RAG industrialisé (hallucinations, hors-sujets, langue, citations, abstentions).
    
    \item Une prise en compte explicite de la température non nulle via des répétitions multiples.
    
    \item Un protocole hybride combinant annotations humaines et LLM-Juge, inspiré de PPI.
    
    \item Un jeu de données complet en français (questions, réponses, sources, annotations humaines et LLM-Juge), anonymisé et accessible publiquement : \url{https://github.com/gmartinonQM/eval-rag}
    
\end{itemize}

Dans cet article, nous retenons la définition d'une hallucination comme étant un élément non-déductible d'une base de connaissances vérifiable \cite{maleki2024ai}. 

\section{Travaux connexes}

L’évaluation des LLMs est cruciale, notamment dans les systèmes à haut risque \cite{weidinger2025toward, zhou2024trustworthiness}. Les benchmarks comme HELM \cite{liang2023holistic} s’appuient sur des jeux de données labellisés, incluant des ensembles spécialisés pour détecter les hallucinations. ANAH \cite{ji-etal-2024-anah} et HaluEval \cite{li-etal-2023-halueval} les caractérisent à la maille phrase dans des contextes généraux, là où TofuEval \cite{tang-etal-2024-tofueval} se focalise sur les résumés de dialogue. RAGTruth \cite{niu-etal-2024-ragtruth} descend à la maille fait et se réfère à des passages du jeu de données MS Marco \cite{conf/nips/NguyenRSGTMD16}, tandis que LLM-OASIS \cite{scire2024truth} s'intéresse aux faits en rapport avec Wikipédia. Notre jeu de données s’inscrit dans la lignée de MEMERAG \cite{blandon2025memerag}, qui évalue les hallucinations et les hors-sujets à l'échelle de la phrase dans un contexte RAG, mais avec un ancrage industriel plus fort dans la due diligence.

Les jeux de données standards servent souvent à entraîner des détecteurs d’hallucinations, mais restent limités pour évaluer un système en contexte réel. Les méthodes basées sur un LLM-Juge \cite{zheng2023judging} ou un comité de LLM-Juges \cite{chern2024towards, jung2025trust}, comme RAGAS \cite{es2024ragas} ou G-Eval \cite{liu-etal-2023-g} permettent une évaluation automatique et scalable. Ces méthodes peuvent être enrichies d'outils tels que Google Search ou des interpréteurs Python, comme dans les approches SAFE \cite{NEURIPS2024_937ae0e8} et FactTool \cite{chern2023factool}. Certaines méthodes exploitent la stochasticité des LLMs pour estimer des scores d'hallucinations, c'est le cas de ChainPoll \cite{friel2023chainpoll} ou des approches  par entropie sémantique \cite{farquhar2024detecting}. D'autres méthodes vont plus loin, et calibrent ces scores pour refléter la probabilité d'occurrence d’une hallucination \cite{valentin2024cost}. Cependant, ces méthodes sont sensibles au prompt, au modèle utilisé, et à la qualité des données disponibles, comme les graphes de connaissance \cite{mountantonakis2023validating}. Les résultats obtenus par les LLM-Juges sont très variables, avec des performances oscillant entre 5\% et 50\% \cite{hong2024hallucinations}.

À l’opposé du spectre, les annotations humaines offrent une grande précision mais sont coûteuses et difficilement scalables \cite{min2023factscore}. Des méthodes hybrides comme PPI \cite{doi:10.1126/science.adi6000, angelopoulos2023ppi++} et ASI \cite{10.5555/3692070.3694680} combinent les deux approches. Elles incarnent une théorie des sondages augmentée par LLM. PPI repose sur un échantillon de contrôle aléatoire annoté par des humains et LLM-Juge. Les annotations humaines sont utilisées pour corriger les biais du LLM-Juge, et appliquer cette correction à grande échelle. ASI affine encore cette logique via un échantillonnage adaptatif basé sur l’incertitude du LLM-Juge. PPI a été appliqué à des systèmes de compétition entre LLMs, tels que ChatbotArena \cite{boyeau2024autoeval}, ou à des systèmes RAGs \cite{saad-falcon-etal-2024-ares}. De son côté, ASI a été utilisé pour des systèmes de classification multi-classe basés sur des LLMs \cite{gligoric2024can}. Nos travaux s'inscrivent dans la lignée de ARES \cite{saad-falcon-etal-2024-ares}, en exploitant l'approche PPI dans le contexte industriel de la due diligence.

\section{Système évalué}

\subsection{Contexte industriel}

Dans cet article, nous nous concentrons sur l’évaluation d’Alban, un assistant virtuel développé pour accompagner un fonds d’investissement international gérant plusieurs milliards d’euros d’actifs. Ce type d’acteur mène régulièrement des opérations de due diligence, c’est-à-dire un processus d’analyse approfondie préalable à une décision d’investissement, de fusion ou d’acquisition. L’objectif est d’évaluer de manière rigoureuse la situation financière, juridique, opérationnelle et stratégique d’une entreprise cible, afin d’identifier les risques potentiels, de confirmer la valeur réelle de l’actif, et de négocier au mieux les termes de la transaction. Cette démarche repose sur l’étude de très nombreux documents internes (rapports d’activité, états financiers, contrats, procès-verbaux, audits, etc.) produits par l’entreprise analysée et transmis dans une \textit{data room} dédiée.

Pour un fonds de grande envergure, ce processus représente un effort considérable : plusieurs millions d’euros sont investis à chaque due diligence dans l’analyse documentaire, avec un volume dépassant les 100 000 documents à examiner en quelques semaines seulement. Alban a été conçu pour rationaliser cette phase, en offrant aux équipes de transaction un agent conversationnel capable de répondre rapidement et précisément à leurs questions, à partir du corpus documentaire fourni.

La solution repose sur GPT-4, utilisé avec une température non nulle, et est aujourd’hui en production avec plus de 50 utilisateurs actifs.

\subsection{Système RAG}

Alban est un système RAG (Retrieval-Augmented Generation) de type "Advanced" \cite{gao2023retrieval}. Il commence par ingérer le patrimoine documentaire de l’entreprise, tous formats confondus (PDF, PowerPoint, Word, Excel), en le découpant en segments appelés sources. Chaque source est enrichie de métadonnées (titre, page) et vectorisée à l’aide d’un moteur d’embeddings, puis stockée dans une base vectorielle.

Lors d’une requête, le prompt utilisateur est reformulé, vectorisé et comparé à la base pour extraire les 15 sources les plus proches. Ces sources, combinées à la requête, sont fournies au LLM via un système de prompt en anglais pour générer une réponse pertinente.

La figure \ref{fig:datacorp_evaluation} illustre un exemple d'échange avec Alban.

\begin{figure}[htbp]
   \centering
   \fbox{
   \begin{minipage}{0.95\textwidth}
       \textbf{Question} \\
       Comment DataCorp évalue-t-elle la satisfaction et l'expérience des salariés ? \\
       \textbf{Réponse} \\
       DataCorp utilise des enquêtes de satisfaction pour obtenir des commentaires sur l'engagement, le moral et la satisfaction des employés au travail. Ces enquêtes sont menées dans le cadre des dispositifs convenus avec ComeToMyCorp, une entreprise spécialisée dans l'évaluation de la satisfaction des employés [\textasciicircum 5f7cce\textasciicircum]. Il est à noter que la direction de DataCorp a également mis en place des actions pour répondre aux commentaires laissés sur la plateforme JobReview, afin de gérer l'e-réputation de l'entreprise [\textasciicircum 4ca822\textasciicircum][\textasciicircum 63fadb\textasciicircum].
   \end{minipage}
   }
   \caption{Exemple de réponse générée par le système Alban. Les identifiants entre crochets correspondents aux identifiants des sources sur lesquelles le système base sa réponse.}
   \label{fig:datacorp_evaluation}
\end{figure}

\section{Points de défaillance du système}

En inspectant manuellement les réponses du système Alban, plusieurs points de défaillance ont été recensés :

\begin{itemize}
    \item \textbf{Réponses stochastiques} : Le système peut donner une réponse différente si la même question lui est posée plusieurs fois.
   
   \item \textbf{Langue erronée} : Le système peut répondre en anglais, même si l'utilisateur écrit en français.

   \item \textbf{Réponses inattendues} : Le système peut répondre à des questions pour lesquelles aucune donnée n'est disponible.
   
   \item \textbf{Citations défaillantes} : Le modèle peut mal restituer les identifiants des sources, par exemple en citant \texttt{4ca823} au lieu de \texttt{4ca822}.
   
   \item \textbf{Hors-sujets} : Le système peut fournir des informations sans rapport avec la question posée.
   
   \item \textbf{Langue de bois} : Le système peut inclure des éléments non-assertifs, par exemple de pure politesse, ou des conjectures non vérifiables.
   
   \item \textbf{Hallucinations} : Le système peut inclure des éléments qui ne sont pas déductibles des sources citées.
   
   \item \textbf{Réponses partielles} : Le système peut fournir une réponse partielle, même lorsqu'il dispose de tous les éléments nécessaires dans les sources.
   
\end{itemize}

Dans le contexte de la due diligence, ces points de défaillance ont des impacts qui peuvent aller de la simple perte de temps (langue erronée, hors-sujets, langue de bois), à des décisions stratégiques d'investissement faussées (hallucinations, réponses partielles).

Dans la suite, nous traitons les risques à la maille réponse, sauf pour les \textbf{Citations défaillantes} et les \textbf{Hallucinations} qui sont traitées à la maille phrase. Par ailleurs, nous choisissons de ne pas traiter le problème des \textbf{Réponses partielles}, dont l’évaluation nécessite une expertise métier très spécialisée, ni de la \textbf{Langue de bois}, qui est une notion relativement subjective et moins critique.

\section{Construction du jeu de données}

\paragraph{Jeu de questions} Nous avons reconstitué le patrimoine documentaire de l’entreprise cible, DataCorp (nom modifié), un cabinet de conseil en informatique, en intégrant 300 documents dans notre base. En parallèle, un questionnaire de due diligence de 121 questions a été élaboré avec des experts métiers, puis classé par thème et par niveau de difficulté (voir Table \ref{tab:complexite_themes}). Certaines questions, inappropriées au contexte, ont été conservées pour évaluer la capacité du système à s’abstenir de répondre \footnote{Le jeu de données anonymisé est disponible à cette adresse : \url{https://github.com/gmartinonQM/eval-rag}. Le détail de la procédure d'anonymisation est présenté en Section \ref{sec:materiel_supplementaire}.}.

\begin{table}[h]
   \centering
   \footnotesize
   \begin{tabular}{|l|c|c|c|c|}
       \hline
       \textbf{Thème/Difficulté} & \textbf{Simple} & \textbf{Intermédiaire} & \textbf{Difficile} & \textbf{Inapproprié} \\
       \hline
       Finance & 16 & 16 & 16 & 15 \\
       RH & 7 & 7 & 7 & 3 \\
       IT & 10 & 10 & 10 & 4 \\
       \hline
   \end{tabular}
   \caption{Répartition des questions de notre jeu de données par thème et par niveau de difficulté.}
   \label{tab:complexite_themes}
\end{table}

\paragraph{Jeu de réponses} Pour chaque question, nous avons généré 20 réponses distinctes, en conservant le même libellé et en initiant une nouvelle conversation à chaque fois. Cette procédure permet de prendre en compte les \textbf{réponses stochastiques} à température non nulle \footnote{Le choix de 20 répétitions repose sur une étude exploratoire de répétabilité : nous avons observé que le taux de véracité par question se stabilisait seulement à partir de ce seuil. Un budget plus important aurait permis de générer davantage de répétitions pour obtenir des métriques robustes à l'échelle de chaque question. En pratique, pour limiter les coûts de calcul et d’annotation, nous restituons les résultats agrégés à la maille thème, et considérons que 20 réponses suffisent déjà à capturer une variabilité suffisante du système à cette maille.}.

\begin{figure}[htbp] 
\begin{center} 
\includegraphics[width=0.7\textwidth]{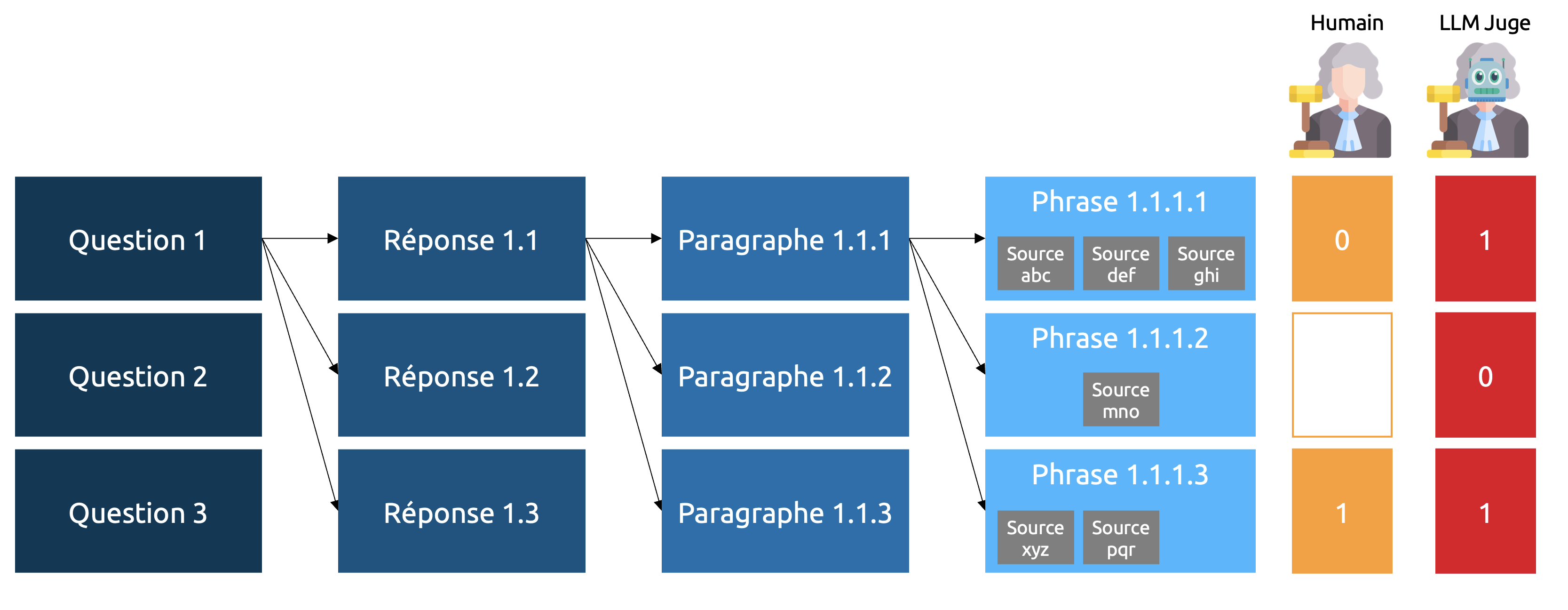}
\end{center} 
\caption{Protocole d'évaluation du système RAG. Pour chaque question, 20 réponses différentes sont générées. Les réponses sont découpées en phrases, chacune avec ses sources. Un annotateur humain et LLM-Juge évaluent les évaluent avec une notation binaire (0 ou 1). Le juge humain n'évalue qu'un échantillon aléatoire, tandis que le LLM-Juge évalue l'intégralité.}
\label{fig:database}
\end{figure}

\paragraph{Extraction de phrases} Afin de mieux quantifier et localiser les hallucinations, nous découpons chaque réponse en phrases, comme indiqué en Figure \ref{fig:database}.

\paragraph{Stratégie d'échantillonnage} Pour les réponses comme pour les phrases, nous sélectionnons les observations à annoter par un plan de sondage stratifié. Dans un premier temps, nous effectuons un embedding de toutes les observations avec le moteur \texttt{text-embedding-ada-002} d'OpenAI. Ensuite, nous appliquons un algorithme de clustering K-Means à ces embeddings au sein de chaque thème. Enfin, nous sélectionnons aléatoirement trois observations par cluster. Le nombre de clusters K est déterminé par le budget alloué à l'annotation humaine.

\paragraph{Protocole d'annotation} Les annotations, humaines ou produites par un LLM-Juge, permettent de détecter les hors-sujets et les hallucinations. Les consignes, identiques pour les deux types d’annotateurs, sont formalisées sous forme de prompt. Trois annotateurs humains, de niveaux d’expertise croissants, interviennent successivement : le second relit le premier, et un troisième tranche en cas de désaccord persistant. Tous sont coauteurs du présent article. Un exemple d’annotation est présenté en Section \ref{sec:materiel_supplementaire}.

\paragraph{LLM-Juge} Dans cet article, le LLM-Juge est GPT-4o (\texttt{gpt-4o-2024-08-06}) utilisé à température nulle.

Le jeu de données ainsi constitué ne saurait être considéré comme un \textit{gold standard} : obtenir une réponse idéale à chaque question impliquerait un coût humain bien supérieur. C’est précisément tout l’intérêt de notre approche : permettre une évaluation à bas coût d’un système génératif, tout en s’appuyant sur une revue humaine experte.

\section{Protocole d'évaluation}

\subsection{Métriques}

Pour chaque risque identifié, nous appliquons des métriques de performance spécifiques.

\paragraph{Taux de langue correcte (maille réponse)} Nous mesurons la proportion de réponses rédigées en français, sachant que toutes les questions sont posées dans cette langue et que nous générons une vingtaine de réponses par question \footnote{Pour détecter la langue des réponses, nous utilisons le package Python \texttt{langdetect}, qui n'a montré aucune erreur sur un échantillon de 1 000 réponses vérifiées manuellement.}.

\paragraph{Taux de réponse  (maille réponse)} Nous calculons la proportion de réponses qui ne sont pas des abstentions manifestes et qui citent explicitement des sources. Nous employons des expressions régulières (REGEX) pour détecter la présence de citations dans le texte.

\paragraph{Taux de citations fonctionnelles  (maille phrase)} Nous calculons la proportion de phrases où tous les IDs de sources citées correspondent aux IDs fournis par le moteur de recherche.

\paragraph{Pertinence (maille réponse)} Nous estimons la proportion de réponses exemptes de tout contenu hors sujet. Nous introduisons un label binaire : "0" si la réponse contient au moins un hors sujet, "1" si la réponse ne contient aucun hors sujet.

\paragraph{Véracité (maille phrase)} Nous évaluons la proportion de phrases dont toutes les affirmations sont déductibles du texte. Nous introduisons un label binaire : "0" la phrase contient au moins un élément non déductible des sources citées (hallucination), "1" la phrase est entièrement déductible des sources citées. Les phrases qui relèvent de la langue de bois sont annotées "1", pour ne pas introduire de confusion avec la notion d'hallucination.

Les métriques \textbf{taux de langue correcte}, \textbf{taux de citations fonctionnelles}, et \textbf{taux de réponse} peuvent être mesurées de manière programmatique et sans ambiguïté. En revanche, les métriques de \textbf{pertinence} et de \textbf{véracité} requièrent du raisonnement, que nous traitons par des annotations humaines ou LLM-Juge. Les prompts utilisés à cette fin sont présentés en Section \ref{sec:materiel_supplementaire}.

\subsection{Prediction-Powered Inference}

Pour obtenir des intervalles de confiance sur ces métriques, nous appliquons la méthode PPI++ \cite{angelopoulos2023ppi++}. Cette méthode vise à fournir une estimation fiable de ces métriques en combinant les annotations humaines sur un échantillon sélectionné avec les annotations réalisées par le LLM-Juge sur l'ensemble du jeu de données. Nous donnons plus de détails sur la méthode en Section \ref{sec:materiel_supplementaire}. 

\section{Résultats}

\begin{figure}[htbp]
   \centering
   \begin{minipage}{0.5\textwidth}
       \centering
       \includegraphics[width=\textwidth]{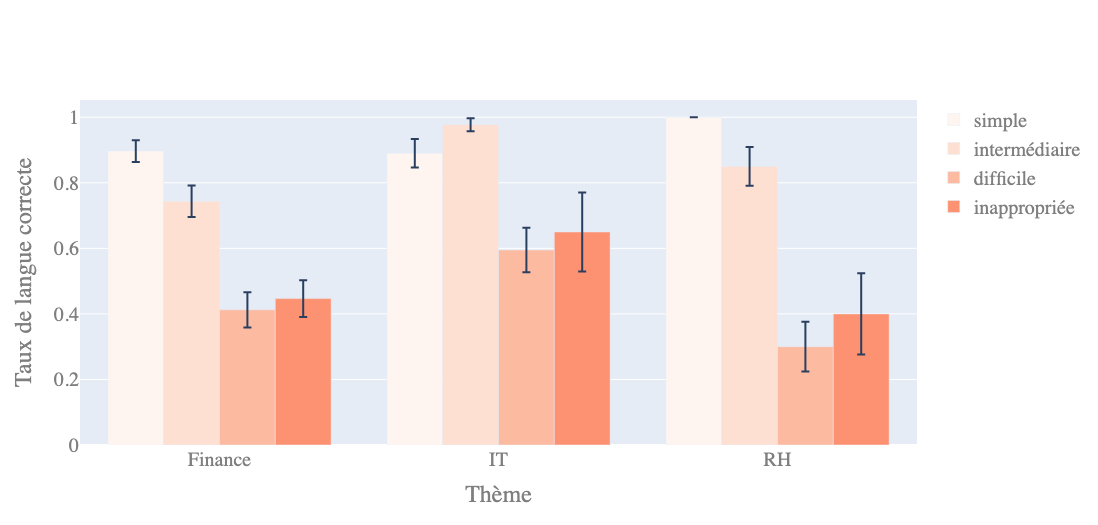}
   \end{minipage}\hfill
   \begin{minipage}{0.5\textwidth}
       \centering
       \includegraphics[width=\textwidth]{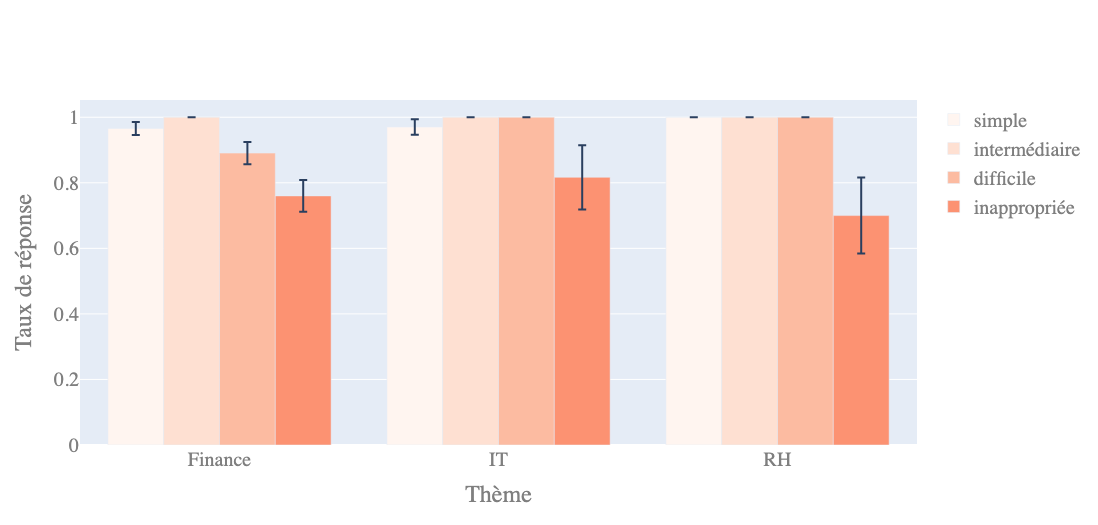}
   \end{minipage}

   \vspace{0.5cm}

   \begin{minipage}{0.5\textwidth}
       \centering
       \includegraphics[width=\textwidth]{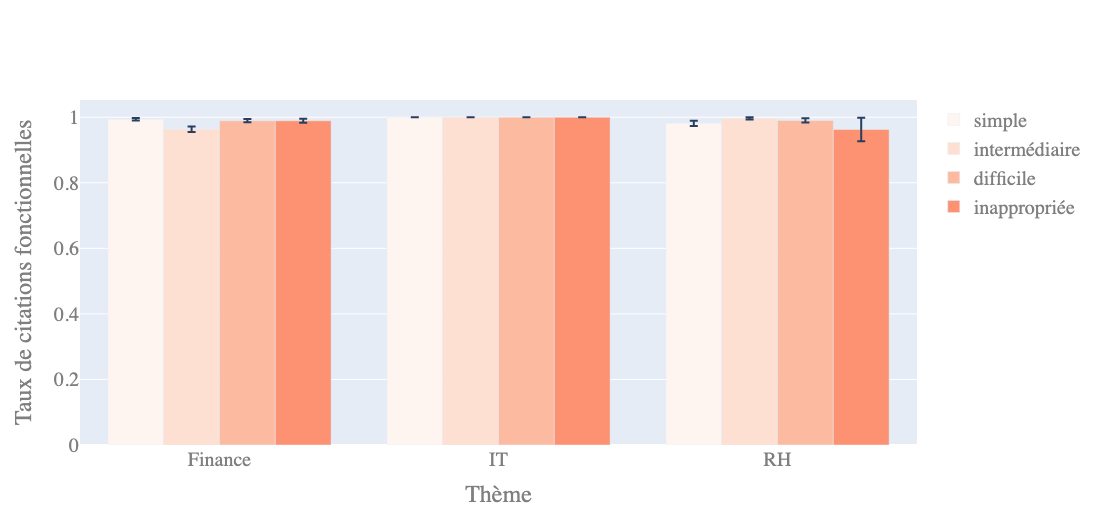}
   \end{minipage}
   \caption{Métriques calculées automatiquement par thème et par niveau de difficulté des questions. En haut à gauche : taux de réponses en français. En haut à droite : taux de réponses effectives. En bas : taux de citations correctes, indiquant si l'identifiant de la source citée est non corrompu. Les barres d'erreurs correspondent aux intervalles de confiance de Wald à 95\%.}
   \label{fig:metric_automated}
\end{figure}

La Figure \ref{fig:metric_automated} présente les métriques calculées automatiquement (langue correcte, réponse effective et citation fonctionnelle) par thématique et niveau de difficulté des questions.

On observe une nette dégradation des performances lorsque la complexité des questions augmente. Le système tend alors à s’abstenir de répondre et à basculer en anglais, un comportement également observé pour les questions inappropriées, pourtant censées rester sans réponse. Cela suggère que face à une faible confiance (mesuré par exemple par sa perplexité \cite{Jelinek1977PerplexityaMO}), le modèle opte pour l’anglais.
Par ailleurs, le système fournit une réponse même lorsqu’aucune n’est attendue. Enfin, bien qu’il restitue correctement les identifiants de sources dans 99\% des cas, les erreurs résiduelles peuvent nuire à son exploitation.

\begin{figure}[htbp]
   \centering
   \begin{minipage}{0.5\textwidth}
       \centering
       \includegraphics[width=\textwidth]{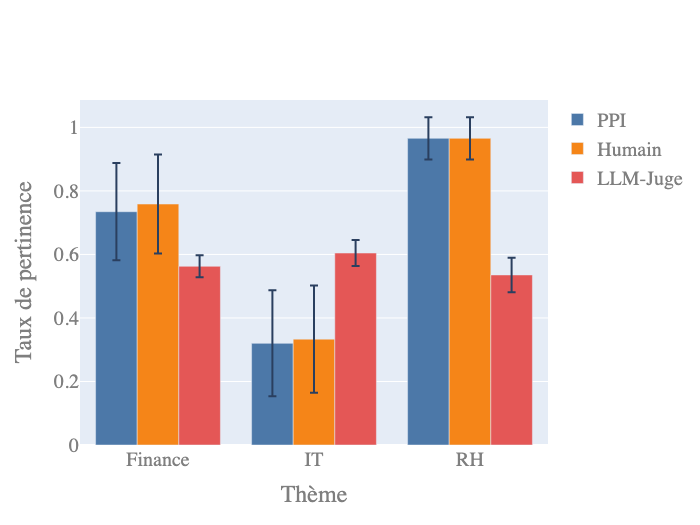}
   \end{minipage}\hfill
   \begin{minipage}{0.5\textwidth}
       \centering
       \includegraphics[width=\textwidth]{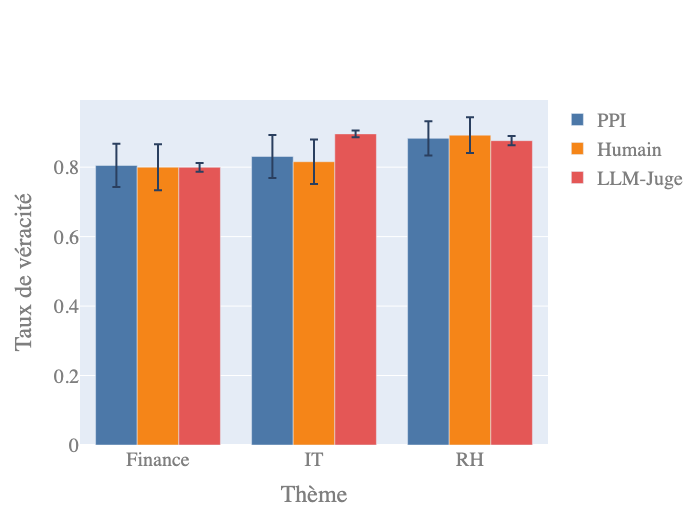}
   \end{minipage}
   \caption{Métriques calculées par annotations humaine et LLM-Juge par thème. A gauche : taux de pertinence. A droite : taux de véracité. Les barres d'erreurs pour "Humain" et "LLM" correspondent aux intervalles de confiance de Wald à 95\%, tandis que les barres d'erreurs pour "PPI" sont directement issues de la méthode PPI.}
   \label{fig:metric_annotated}
\end{figure}

La Figure \ref{fig:metric_annotated} présente les métriques nécessitant un raisonnement (pertinence et véracité) et compare les performances des trois méthodes d’évaluation : annotations humaines, LLM-Juge et PPI.

Les annotations humaines révèlent une forte variabilité de pertinence selon les thématiques, avec des résultats particulièrement faibles en IT (32\% de réponses entièrement pertinentes). La véracité reste relativement élevée, entre 80\% et 88\% des phrases. Une analyse manuelle révèle que les 12 à 20\% de phrases contenant une hallucination sont bien réparties dans quasiment toutes les réponses.

Comparé à l’humain, le LLM-Juge surestime nettement la pertinence, notamment en IT (écart d’un facteur 2), tandis que les écarts de véracité restent plus modérés (jusqu’à 6\%).

Les estimations PPI, très proches de celles obtenues par annotations humaines, montrent que dans cette étude, les annotations du LLM-Juge apportent peu d’information utile.

Ce résultat s’explique par le faible taux d’accord entre annotations humaines et LLM-Juge sur l’échantillon de contrôle. Comme le montrent les Tables \ref{tab:pertinence} et \ref{tab:veracite}, l’accord observé frôle parfois le niveau aléatoire, signalant une faible exploitabilité du LLM-Juge. Dans ces conditions, l’incertitude obtenue par PPI est proche de celle d’un simple sondage exploitant les annotations humaines, ce qui rend la contribution du LLM-Juge marginale. Ces constats rejoignent ceux de \cite{gligoric2024can} sur d’autres jeux publics. Une analyse de sensibilité à l’accord humain/LLM-Juge est proposée en Section \ref{sec:materiel_supplementaire}.

\begin{table}[htbp]
\centering
\footnotesize
\begin{tabular}{|l|c|c|c|c|c|}
\hline
\textbf{Thème} & 
\textbf{Accord aléatoire} & 
\textbf{Accord observé} & 
\textbf{Ann. hum.} & 
\textbf{Ann. hum. eff.} & 
\textbf{Ann. LLM-Juge} \\
\hline
Finance & 0.62 & 0.69 & 29 & 31.98 & 791 \\
IT & 0.43 & 0.50 & 30 & 30.04 & 551 \\
RH & 0.61 & 0.59 & 29 & 29.00 & 325 \\
\hline
\end{tabular}
\caption{Gain apporté par la combinaison du LLM-Juge et de PPI pour l’évaluation de la pertinence.
L’accord aléatoire correspond au taux attendu si les annotations humaines et LLM-Juge étaient générées indépendamment selon une loi de Bernoulli ; il sert de référence de pire cas. L’accord observé indique la proportion réelle d’annotations concordantes sur l’échantillon de contrôle. Ann. hum. et Ann. LLM-Juge désignent respectivement le volume d’annotations humaines sur l’échantillon et celui, complet, du LLM-Juge. Ann. hum. eff. représente la taille effective des annotations humaines, qui aurait été nécessaire avec un sondage classique  pour atteindre la même incertitude que celle observée avec PPI.}
\label{tab:pertinence}
\end{table}

\begin{table}[htbp]
\centering
\footnotesize
\begin{tabular}{|l|c|c|c|c|c|}
\hline
\textbf{Thème} & 
\textbf{Accord aléatoire} & 
\textbf{Accord observé} & 
\textbf{Ann. hum.} & 
\textbf{Ann. hum. eff.} & 
\textbf{Ann. LLM-Juge} \\
\hline
Finance & 0.67 & 0.79 & 140 & 155.59 & 3985 \\
IT & 0.73 & 0.80 & 141 & 141.57 & 3799 \\
RH & 0.82 & 0.88 & 139 & 162.36 & 2408 \\
\hline
\end{tabular}
\caption{Gain apporté par l’utilisation conjointe du LLM-Juge et de PPI pour l’évaluation de la véracité.}
\label{tab:veracite}
\end{table}

\section{Conclusion}

Dans cet article, nous avons évalué un système RAG dans un contexte industriel de due diligence, en analysant ses défaillances par domaine métier et en combinant plusieurs méthodes d’évaluation. Nos résultats montrent que, malgré leur potentiel pour automatiser l’analyse documentaire, ces systèmes soulèvent des enjeux critiques de fiabilité, notamment en termes de pertinence et de véracité.

Le protocole proposé, fondé sur des métriques automatiques, des annotations humaines et des évaluations par LLM-Juge, permet une mesure scalable et réplicable à chaque itération du système. Les deux facteurs déterminants pour réduire l’incertitude sont le volume d’annotations humaines et la qualité du LLM-Juge. Or, la méthode PPI ne devient réellement avantageuse que si l'accord entre humain et LLM-Juge est excellent, ce qui suppose un travail rigoureux de conception de prompt, parfois plus coûteux en temps qu’une session supplémentaire d'annotation humaine.

Intégrée dès les premières phases de développement, cette approche permet néammoins d'alimenter une boucle de rétroaction continue : affiner le prompt du LLM-Juge, ajuster les consignes d’annotation ou enrichir un corpus pour du \textit{fine-tuning} du LLM-Juge. Ce cycle de capitalisation progressive constitue, selon nous, une voie prometteuse pour fiabiliser durablement l’évaluation des systèmes génératifs.

\section{Limitations et directions futures}

Le protocole proposé ouvre plusieurs perspectives d'amélioration.

D’abord, la granularité d’annotation pourrait être affinée : au lieu d’évaluer la véracité au niveau des phrases, on pourrait cibler des plages de mots, comme dans les jeux de données ANAH \cite{ji-etal-2024-anah} ou HaluEval \cite{li-etal-2023-halueval}.

Ensuite, l'échantillonnage utilisé ici repose sur des embeddings généralistes. L’adoption d’un moteur d’embedding adapté au contexte de la due diligence pourrait produire un clustering plus pertinent, et donc réduire l’incertitude via un échantillonnage mieux stratifié.

Notre étude s’est par ailleurs limitée à un seul LLM-Juge (GPT-4o). En comparant plusieurs modèles et prompts, il serait possible d’identifier les configurations les plus fiables, en s’appuyant sur l’incertitude calculée par PPI comme indicateur de performance. Une telle sélection de couple LLM/prompt nécessiterait toutefois un jeu d’annotations de validation pour éviter tout sur-apprentissage.

Le protocole pourrait aussi être étendu à des variantes méthodologiques, telles que ASI, mieux adaptées aux sondages stratifiés que ne l’est PPI, fondé sur un plan de sondage simple.

Enfin, une piste supplémentaire consisterait à intégrer la langue de bois comme nouvelle dimension d’évaluation, à l’aide d’un protocole dédié.

\textbf{N.B.} Cet article a été rédigé avec l'assistance de GPT-4o.

\section{Matériel supplémentaire}
\label{sec:materiel_supplementaire}

\subsection{Procédure d'anonymisation des données}

Les données ont été systématiquement anonymisées selon le protocole suivant :
\begin{itemize}
    \item Remplacement de toutes les adresses postales et noms de ville
    \item Substitution de tous les prénoms et noms de famille
    \item Changement des noms d’entreprise
    \item Réécriture du vocabulaire spécifique à l’entreprise cible
    \item Remplacement des adresses e-mail par \texttt{johndoe@company.com}
    \item Décalage temporel de toutes les dates vers le passé
    \item Substitution des liens web par \texttt{https://example.com}
    \item Modification de tous les nombres, pourcentages et montants financiers
    \item Remplacement des numéros de téléphone par \texttt{01 23 45 67 89}
    \item Remplacement des numéros SIREN par \texttt{123 456 789}
    \item Remplacement des numéros de compte bancaire par \texttt{123456789}
\end{itemize}

Ce protocole d’anonymisation vise à garantir la protection de l’identité de l’entreprise cible, de l’entreprise acquérante ainsi que de toutes les personnes mentionnées dans les documents analysés.

\subsection{Exemples d'annotations}

\begin{figure}[htbp] 
\begin{center} 
\includegraphics[width=\textwidth]{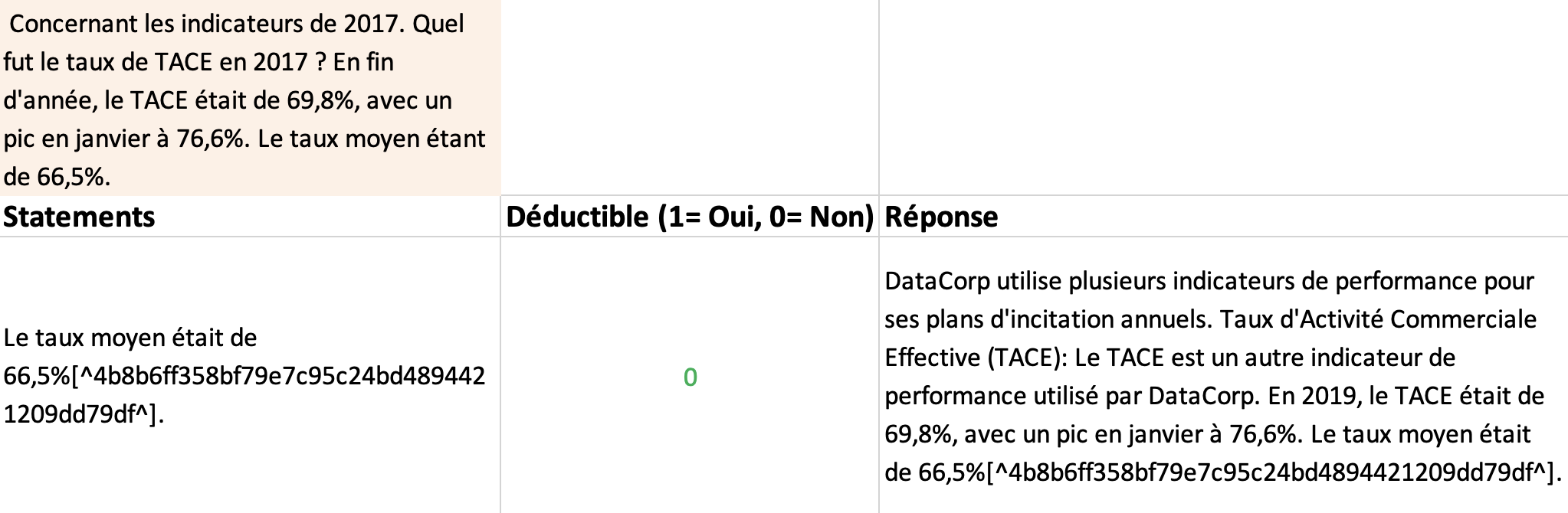}
\end{center} 
\caption{Exemple d'annotation d'une hallucination. L'énoncé présenté ("Le taux moyen était de 66,5\%") est incorrectement déduit de la source (en haut à gauche) : la source décrit l'année 2017, quand la réponse décrit l'année 2019. L'annotation attribue donc le label \textbf{0} (non déductible).}
\label{fig:annotation_veracite}
\end{figure}

La figure \ref{fig:annotation_veracite} illustre une capture d'écran d'un fichier excel ayant permis d'annoter une phrase générée par le système évalué Alban. En l'occurrence, l'exemple illustre une hallucination au sens où la réponse et la source se situent à des dates différentes.

\subsection{Prompts du LLM-Juge}

Le prompt du LLM-Juge sur la véracité est présenté en Figure \ref{fig:prompt_template_faithfulness}.

\begin{figure}[htbp]
   \centering
   \fbox{
        \begin{minipage}{0.95\textwidth}
        \small
        \textbf{Rôle du LLM-Juge} :\\
        Votre mission est d'analyser la véracité des phrases fournies par rapport à des documents de référence. Dès qu'une information de la phrase est non-déductible des sources, alors toute la phrase doit être classée comme non-déductible.
        
        \vspace{0.5em}
        Les données qui vous seront fournies sont :
        \begin{itemize}
            \item \textbf{Sources} : Les documents de référence.
            \item \textbf{Paragraphe} : Le paragraphe contextualisant la phrase.
            \item \textbf{Phrases} : La phrase dont vous devez évaluer le caractère déductible.
        \end{itemize}
        
        \vspace{0.5em}
        \textbf{Instructions d'évaluation} :\\
        Adoptez l'approche suivante dans votre évaluation :
        \begin{enumerate}
            \item Lire les sources en entier.
            \item Lire le paragraphe en entier.
            \item Lire les phrases en entier.
            \item Pour chaque phrase, verbaliser un raisonnement pas à pas sur le caractère déductible des informations de la phrase par rapport aux sources.
            \item Si les informations de la phrase sont insuffisantes ou ambiguës, vous pouvez utiliser vos connaissances du monde pour déterminer si toutes ces informations sont réellement déductibles des sources.
            \item En déduire une évaluation finale pour chaque phrase.
        \end{enumerate}
        
        \vspace{0.5em}
        Chaque évaluation est un label :
        \begin{itemize}
            \item \textbf{0} : Au moins une information n'est pas déductible des sources.
            \item \textbf{1} : Toutes les informations sont déductibles des sources.
        \end{itemize}
        
        \vspace{0.5em}
        \textbf{Format des réponses} :\\
        Vous devez renvoyer uniquement un JSON dans la structure suivante :
        \textit{json schema}
        
        Important : Le contenu renvoyé doit être un JSON strictement valide, sans texte supplémentaire, sans explication ni commentaire, directement parsable et la clé « verdicts » devant correspondre à une liste d'objets JSON.
        
        \vspace{0.5em}
        \textbf{Exemples} :
        
        Exemple de sources :
        \textit{example input}

        Exemple de paragraphes :
        \textit{example paragraph}
        
        Exemple de phrase :
        \textit{example statements}
        
        Exemple de réponse au format JSON : :
        \textit{example output}
        
        \vspace{0.5em}
        \textbf{Voici les données} :
        
        Sources :
        \textit{sources}

        Paragraphes :
        \textit{paragraphes}
        
        Phrases :
        \textit{phrases}
        
        JSON :
        \end{minipage}
   }
   \caption{Prompt du LLM-Juge pour la véracité. Le LLM lit les sources et les phrases, applique un raisonnement étape par étape, puis attribue un label (déductible, non-déductible) en suivant un protocole strict. Le résultat final est structuré sous forme d'un fichier JSON.}
   \label{fig:prompt_template_faithfulness}
\end{figure}

Le prompt du LLM-Juge sur la pertinence est présenté en Figure \ref{fig:prompt_template_relevancy}.

\begin{figure}[htbp]
   \centering
   \fbox{
        \begin{minipage}{0.95\textwidth}
        \small
        \textbf{Rôle du LLM-Juge} :\\
        Votre mission est d’évaluer la pertinence d’une réponse par rapport à une question posée. Une réponse est dite pertinente si elle :
        \begin{itemize}
            \item répond à la question posée,
            \item n'introduit pas d'éléments hors sujet,
        \end{itemize}

        Il ne s’agit pas ici d’évaluer la véracité de la réponse, ni son objectivité.
        
        \vspace{0.5em}
        Les données qui vous seront fournies sont :
        \begin{itemize}
            \item \textbf{Question} : La question posée.
            \item \textbf{Réponse} : La réponse à évaluer.
        \end{itemize}
        
        \vspace{0.5em}
        \textbf{Instructions d'évaluation} :\\
        Adoptez l'approche suivante dans votre évaluation :
        \begin{enumerate}
            \item Lire attentivement la question.
            \item Lire attentivement la réponse.
            \item Lire les phrases en entier.
            \item Pour chaque phrase, verbaliser un raisonnement pas à pas sur le caractère pertinent des informations de la réponse.
            \item En déduire une évaluation finale pour chaque phrase.
        \end{enumerate}
        
        \vspace{0.5em}
        Chaque évaluation est un label :
        \begin{itemize}
            \item \textbf{0} : La réponse est partiellement pertinente et contient au moins un hors sujet.
            \item \textbf{1} : La réponse est totalement pertinente.
        \end{itemize}
        
        \vspace{0.5em}
        \textbf{Format des réponses} :\\
        Vous devez renvoyer uniquement un JSON dans la structure suivante :
        \textit{json schema}
        
        Important : Le contenu renvoyé doit être un JSON strictement valide, sans texte supplémentaire, sans explication ni commentaire, directement parsable et la clé « verdicts » devant correspondre à une liste d'objets JSON.
        
        \vspace{0.5em}
        \textbf{Exemples} :
        
        Voici un ensemble d'exemples de phrases et de leurs évaluations.

        Exemple de question:
        \textit{example question}
        
        Exemple de réponse:
        \textit{example answer}
        
        Exemple de réponse au format JSON : :
        \textit{example output}
        
        \vspace{0.5em}
        \textbf{Voici les données} :
        
        Question :
        \textit{question}
        
        Réponse :
        \textit{answer}
        
        JSON :
        \end{minipage}
   }
   \caption{Prompt du LLM-Juge pour la pertinence. Le LLM lit la question et la réponse, applique un raisonnement étape par étape, puis attribue un label (totalement pertinent, contient au moins un hors-sujet) en suivant un protocole strict. Le résultat final est structuré sous forme d'un fichier JSON.}
   \label{fig:prompt_template_relevancy}
\end{figure}

\subsection{PPI}

En notant $n$ la taille de l'échantillon annoté par l'homme,  $X_j$ les réponses associées, $Y_j$ les annotations humaines correspondantes, $N$ la taille de l'échantillon uniquement annoté par le LLM-Juge, $\tilde{X}_i$ les réponses associées, et $f(.)$ l'annotation LLM-Juge, on peut estimer de manière non biaisée la moyenne des $Y_i$ sur l'ensemble des données, notée $\hat{\theta}$, par l'expression suivante :

\begin{equation}
    \hat{\theta} = \frac{1}{N} \sum_{i=1}^{N} \lambda f(\tilde{X}_i) - \frac{1}{n} \sum_{j=1}^{n} \left( \lambda f(X_j) - Y_j\right)
\end{equation}

En première approche, le paramètre $\lambda$ peut être considéré égal à 1. Cependant, comme indiqué dans \cite{angelopoulos2023ppi++}, si les évaluations du LLM-Juge sont de mauvaise qualité, les inclure dans l'estimation peut élargir l'intervalle de confiance par rapport à ce qui serait obtenu avec un sondage simple basé uniquement sur les $Y_j$. Pour éviter ce problème, on introduit le paramètre $\lambda \in [0, 1]$, qui peut être optimisé pour garantir des estimations aussi précises, voire meilleures, que celles obtenues par un sondage simple, indépendamment de la qualité des annotations du LLM-Juge ; c'est le \textit{power tuning}. L'expression analytique du paramètre $\lambda$ optimal dépend des observations faites sur le jeu de données.

L'intervalle de confiance au niveau $1 - \alpha$ sur $\hat{\theta}$ s'obtient, lorsque $\lambda=1$, par : 

\begin{equation}
    C_{1-\alpha} = \hat{\theta} \pm z_{1-\alpha/2} \sqrt{\left(\frac{\hat{\sigma}_f^2}{N} + \frac{\hat{\sigma}_{f-Y}^2}{n}\right)}
\end{equation}

où $z_{1-\alpha/2}$ est le quantile $1 - \alpha / 2$ de la distribution normale standardisée,  $\hat{\sigma}_f^2$ et $\hat{\sigma}_{f-Y}^2$ représentent les estimations de variance de $f(\tilde{X}_i)$ et $f(X_j) - Y_j$ respectivement. La formule analytique dans le cas où $\lambda \neq 1$ est donnée en Section 6 de \cite{angelopoulos2023ppi++}. Dans cet article, nous utilisons le package python \texttt{ppi\_py} disponible à cette adresse \url{https://github.com/aangelopoulos/ppi_py}, et développé par les auteurs de la méthode.

\subsection{Analyse de sensibilité de PPI à l'accord humain/LLM-Juge}

Ayant pu observer que le gain sur l'incertitude de mesure apporté par le LLM-Juge et PPI était marginal, notamment à cause d'un taux d'accord insuffisant entre les annotations humaines et LLM-Juge sur l'échantillon de contrôle, on peut légitimement se demander comment évolue l'incertitude de PPI avec l'accord humain/LLM-Juge. 

Nous avons mené une simulation reproduisant les caractéristiques observées sur la métrique de véracité en finance (cf. Table \ref{tab:veracite}), tout en faisant varier artificiellement le taux d’accord entre annotations humaines et LLM-Juge. Dans cette simulation, nous avons considéré 140 annotations humaines, 3985 annotations LLM-Juge, et des métriques mesurées par l'homme et le LLM-Juge identiques, toutes deux égales à 0.8. Le code de simulation est disponible en libre accès : \url{https://github.com/gmartinonQM/eval-rag}.

On constate que pour passer d'une incertitude de 7\% (sondage classique avec 140 observations) à une incertitude de 4\%, il faut être capable de concevoir un LLM-Juge qui s'accorde avec l'être humain dans 93\% des cas. Si tel était le cas, l'incertitude de mesure de 4\% ainsi obtenue serait la même qu'un sondage classique utilisant 375 annotations, soit un facteur 2.7 de gagné sur le temps d'annotations humaines. Ces résultats illustrent de manière quantitative à quel point la qualité du LLM-Juge est déterminante pour que la méthode PPI apporte un avantage substantiel en termes de coût et d’efficacité.

\begin{figure}[htbp]
   \centering
   \begin{minipage}{0.5\textwidth}
       \centering
       \includegraphics[width=\textwidth]{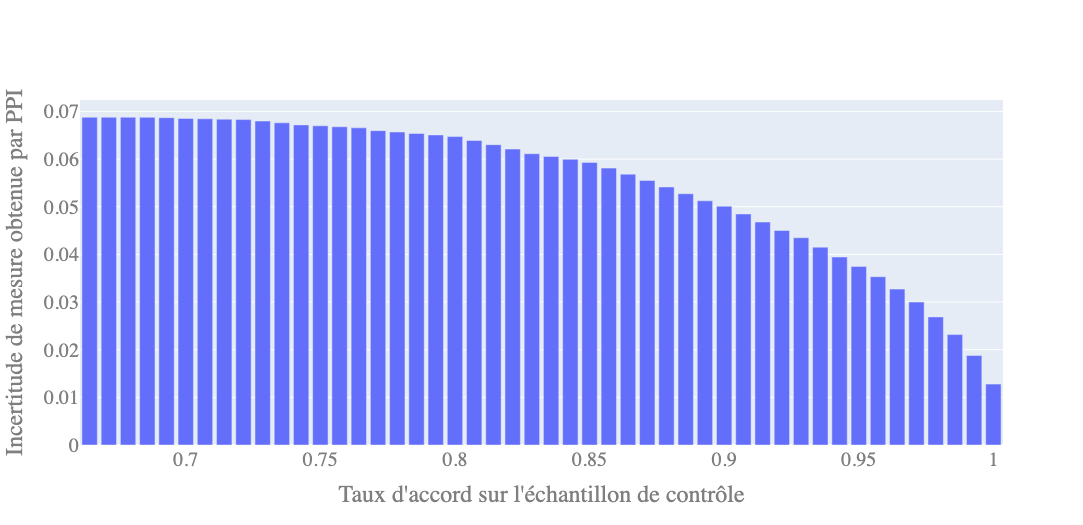}
   \end{minipage}\hfill
   \begin{minipage}{0.5\textwidth}
       \centering
       \includegraphics[width=\textwidth]{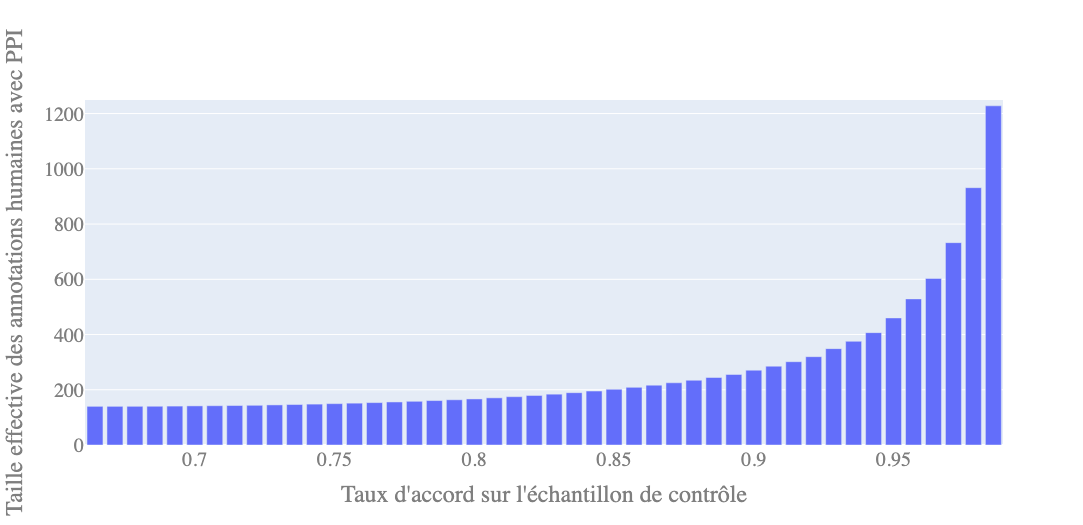}
   \end{minipage}
   \caption{Simulation de résultats obtenus par PPI en faisant varier l'accord humain/LLM-Juge sur l'échantillon de contrôle. A gauche : incertitude de mesure. A droite : taille effective des annotations humaine.}
   \label{fig:metric_ppi_sim}
\end{figure}

\bibliographystyle{coria-taln2025}
\bibliography{biblio}

\end{document}